\documentclass[10pt,twocolumn,letterpaper]{article}

\usepackage{iccv}
\usepackage{times}
\usepackage{epsfig}
\usepackage{graphicx}
\usepackage{amsmath}
\usepackage{amssymb}
\usepackage{subfig}
\usepackage{booktabs}

\usepackage[pagebackref=true,breaklinks=true,letterpaper=true,colorlinks,bookmarks=false]{hyperref}

\iccvfinalcopy 

\def\iccvPaperID{9848} 

\ificcvfinal\fi

\newcommand{\minisection}[1]{\vspace{2mm}\noindent{\textbf{#1}}}
\newcommand{\anh}[1]{{{#1}}}
\newcommand{\nhat}[1]{{#1}}

\begin{document}
\def\mA{\mathcal{A}}
\def\mB{\mathcal{B}}
\def\mC{\mathcal{C}}
\def\mD{\mathcal{D}}
\def\mE{\mathcal{E}}
\def\mF{\mathcal{F}}
\def\mG{\mathcal{G}}
\def\mH{\mathcal{H}}
\def\mI{\mathcal{I}}
\def\mJ{\mathcal{J}}
\def\mK{\mathcal{K}}
\def\mL{\mathcal{L}}
\def\mM{\mathcal{M}}
\def\mN{\mathcal{N}}
\def\mO{\mathcal{O}}
\def\mP{\mathcal{P}}
\def\mQ{\mathcal{Q}}
\def\mR{\mathcal{R}}
\def\mS{\mathcal{S}}
\def\mT{\mathcal{T}}
\def\mU{\mathcal{U}}
\def\mV{\mathcal{V}}
\def\mW{\mathcal{W}}
\def\mX{\mathcal{X}}
\def\mY{\mathcal{Y}}
\def\mZ{\mathcal{Z}} 

\def\bbN{\mathbb{N}} 
\def\bbR{\mathbb{R}} 
\def\bbP{\mathbb{P}} 
\def\bbQ{\mathbb{Q}} 
\def\bbE{\mathbb{E}}

\def\1n{\mathbf{1}_n}
\def\0{\mathbf{0}}
\def\1{\mathbf{1}}

\def\A{{\bf A}}
\def\B{{\bf B}}
\def\C{{\bf C}}
\def\D{{\bf D}}
\def\E{{\bf E}}
\def\F{{\bf F}}
\def\G{{\bf G}}
\def\H{{\bf H}}
\def\I{{\bf I}}
\def\J{{\bf J}}
\def\K{{\bf K}}
\def\L{{\bf L}}
\def\M{{\bf M}}
\def\N{{\bf N}}
\def\O{{\bf O}}
\def\P{{\bf P}}
\def\Q{{\bf Q}}
\def\R{{\bf R}}
\def\S{{\bf S}}
\def\T{{\bf T}}
\def\U{{\bf U}}
\def\V{{\bf V}}
\def\W{{\bf W}}
\def\X{{\bf X}}
\def\Y{{\bf Y}}
\def\Z{{\bf Z}}

\def\a{{\bf a}}
\def\b{{\bf b}}
\def\c{{\bf c}}
\def\d{{\bf d}}
\def\e{{\bf e}}
\def\f{{\bf f}}
\def\g{{\bf g}}
\def\h{{\bf h}}
\def\i{{\bf i}}
\def\j{{\bf j}}
\def\k{{\bf k}}
\def\l{{\bf l}}
\def\m{{\bf m}}
\def\n{{\bf n}}
\def\o{{\bf o}}
\def\p{{\bf p}}
\def\q{{\bf q}}
\def\r{{\bf r}}
\def\s{{\bf s}}
\def\t{{\bf t}}
\def\u{{\bf u}}
\def\v{{\bf v}}
\def\w{{\bf w}}
\def\x{{\bf x}}
\def\y{{\bf y}}
\def\z{{\bf z}}

\def\balpha{\mbox{\boldmath{$\alpha$}}}
\def\bbeta{\mbox{\boldmath{$\beta$}}}
\def\bdelta{\mbox{\boldmath{$\delta$}}}
\def\bgamma{\mbox{\boldmath{$\gamma$}}}
\def\blambda{\mbox{\boldmath{$\lambda$}}}
\def\bsigma{\mbox{\boldmath{$\sigma$}}}
\def\btheta{\mbox{\boldmath{$\theta$}}}
\def\bomega{\mbox{\boldmath{$\omega$}}}
\def\bxi{\mbox{\boldmath{$\xi$}}}
\def\bnu{\mbox{\boldmath{$\nu$}}}                                  
\def\bphi{\mbox{\boldmath{$\phi$}}}
\def\bmu{\mbox{\boldmath{$\mu$}}}

\def\bDelta{\mbox{\boldmath{$\Delta$}}}
\def\bOmega{\mbox{\boldmath{$\Omega$}}}
\def\bPhi{\mbox{\boldmath{$\Phi$}}}
\def\bLambda{\mbox{\boldmath{$\Lambda$}}}
\def\bSigma{\mbox{\boldmath{$\Sigma$}}}
\def\bGamma{\mbox{\boldmath{$\Gamma$}}}
                                  
\newcommand{\myprob}[1]{\mathop{\mathbb{P}}_{#1}}

\newcommand{\myexp}[1]{\mathop{\mathbb{E}}_{#1}}

\newcommand{\mydelta}[1]{1_{#1}}

\newcommand{\myminimum}[1]{\mathop{\textrm{minimum}}_{#1}}
\newcommand{\mymaximum}[1]{\mathop{\textrm{maximum}}_{#1}}    
\newcommand{\mymin}[1]{\mathop{\textrm{minimize}}_{#1}}
\newcommand{\mymax}[1]{\mathop{\textrm{maximize}}_{#1}}
\newcommand{\mymins}[1]{\mathop{\textrm{min.}}_{#1}}
\newcommand{\mymaxs}[1]{\mathop{\textrm{max.}}_{#1}}  
\newcommand{\myargmin}[1]{\mathop{\textrm{argmin}}_{#1}} 
\newcommand{\myargmax}[1]{\mathop{\textrm{argmax}}_{#1}} 
\newcommand{\myst}{\textrm{s.t. }}

\newcommand{\denselist}{\itemsep -1pt}
\newcommand{\sparselist}{\itemsep 1pt}

\definecolor{pink}{rgb}{0.9,0.5,0.5}
\definecolor{purple}{rgb}{0.5, 0.4, 0.8}   
\definecolor{gray}{rgb}{0.3, 0.3, 0.3}
\definecolor{mygreen}{rgb}{0.2, 0.6, 0.2}

\newcommand{\cyan}[1]{\textcolor{cyan}{#1}}
\newcommand{\red}[1]{\textcolor{red}{#1}}  
\newcommand{\blue}[1]{\textcolor{blue}{#1}}
\newcommand{\magenta}[1]{\textcolor{magenta}{#1}}
\newcommand{\pink}[1]{\textcolor{pink}{#1}}
\newcommand{\green}[1]{\textcolor{green}{#1}} 
\newcommand{\gray}[1]{\textcolor{gray}{#1}}    
\newcommand{\mygreen}[1]{\textcolor{mygreen}{#1}}    
\newcommand{\purple}[1]{\textcolor{purple}{#1}}       

\definecolor{greena}{rgb}{0.4, 0.5, 0.1}
\newcommand{\greena}[1]{\textcolor{greena}{#1}}

\definecolor{bluea}{rgb}{0, 0.4, 0.6}
\newcommand{\bluea}[1]{\textcolor{bluea}{#1}}
\definecolor{reda}{rgb}{0.6, 0.2, 0.1}
\newcommand{\reda}[1]{\textcolor{reda}{#1}}

\def\changemargin#1#2{\list{}{\rightmargin#2\leftmargin#1}\item[]}
\let\endchangemargin=\endlist
                                               
\newcommand{\cm}[1]{}

\newcommand{\mhoai}[1]{{\color{magenta}\textbf{[MH: #1]}}}

\newcommand{\mtodo}[1]{{\color{red}$\blacksquare$\textbf{[TODO: #1]}}}
\newcommand{\myheading}[1]{\vspace{1ex}\noindent \textbf{#1}}
\newcommand{\htimesw}[2]{\mbox{$#1$$\times$$#2$}}


\newif\ifshowsolution
\showsolutiontrue

\ifshowsolution  
\newcommand{\Comment}[1]{\paragraph{\bf $\bigstar $ COMMENT:} {\sf #1} \bigskip}
\newcommand{\Solution}[2]{\paragraph{\bf $\bigstar $ SOLUTION:} {\sf #2} }
\newcommand{\Mistake}[2]{\paragraph{\bf $\blacksquare$ COMMON MISTAKE #1:} {\sf #2} \bigskip}
\else
\newcommand{\Solution}[2]{\vspace{#1}}
\fi

\newcommand{\truefalse}{
\begin{enumerate}
	\item True
	\item False
\end{enumerate}
}

\newcommand{\yesno}{
\begin{enumerate}
	\item Yes
	\item No
\end{enumerate}
}

\newcommand{\Sref}[1]{Sec.~\ref{#1}}
\newcommand{\Eref}[1]{Eq.~(\ref{#1})}
\newcommand{\Fref}[1]{Fig.~\ref{#1}}
\newcommand{\Tref}[1]{Table~\ref{#1}}

\title{Toward Realistic Single-View 3D Object Reconstruction \\ with Unsupervised Learning from Multiple Images}

\author{
Long-Nhat Ho$^{1}$ \quad Anh Tuan Tran$^{1,2}$ \quad Quynh Phung$^{1}$ \quad Minh Hoai$^{1,3}$ \\
$^1$VinAI Research, Hanoi, Vietnam,
$^2$VinUniversity, Hanoi, Vietnam,\\
$^3$Stony Brook University, Stony Brook, NY 11790, USA\\
{\tt\small \{v.nhathl,v.anhtt152,v.quynpt29,v.hoainm\}@vinai.io}
}

\makeatletter
\let\@oldmaketitle\@maketitle

\renewcommand{\@maketitle}{\@oldmaketitle
\vspace{-6mm}
\centering
\includegraphics[width=.9\linewidth]{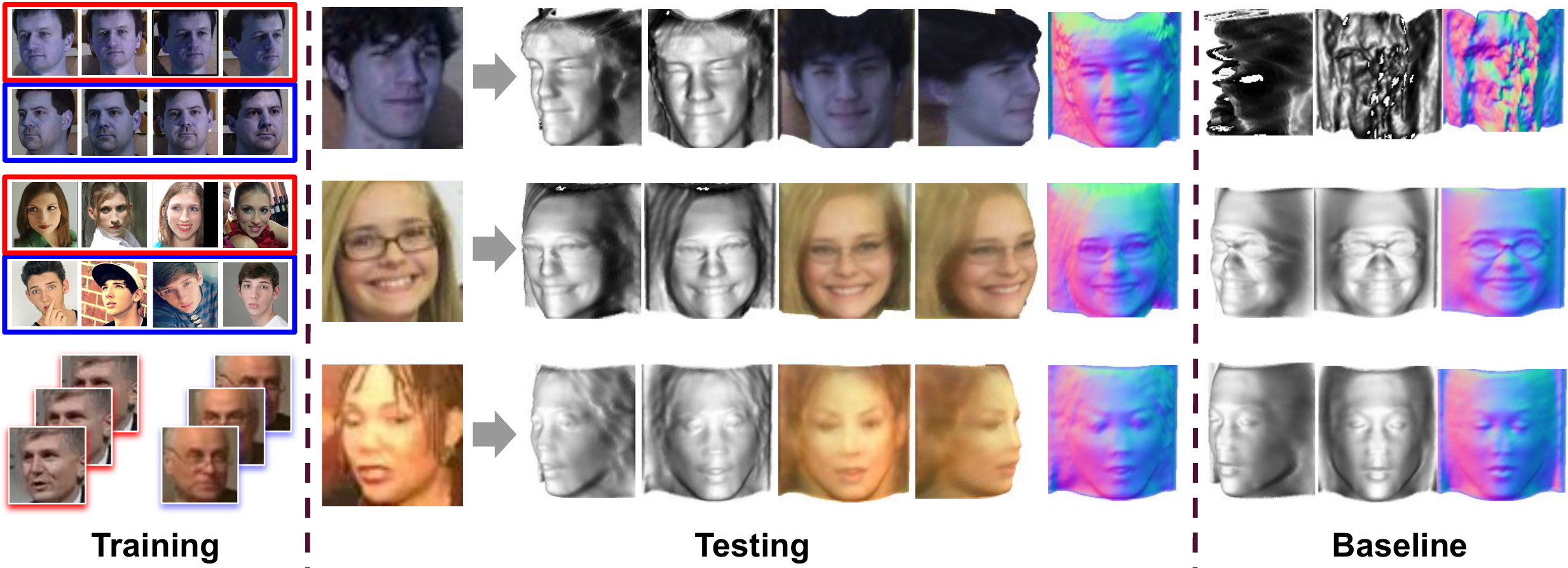}
\vspace{-3mm}
\captionof{figure}{\small We propose a system to learn single-view 3D object reconstruction from multi-image datasets in an unsupervised manner. From top to bottom: Multi-PIE \cite{multiPIE}, CASIA-WebFace \cite{yi2014learning}, and Youtube Faces \cite{wolf2011face} datasets. From left to right: different training data structures, our corresponding 3D reconstructions at test time, and the baseline \cite{wu2020unsupervised} 3D results. For each 3D model, we provide two textureless views, two textured views, and the canonical normal map.} \label{fig:teaser}

\vspace{6mm}
}

\maketitle
\ificcvfinal\thispagestyle{empty}\fi


\begin{abstract}
\vspace{-0.2cm}

Recovering the 3D structure of an object from a single image is a challenging task due to its ill-posed nature. One approach is to utilize the plentiful photos of the same object category to learn a strong 3D shape prior for the object. This approach has successfully been demonstrated by a recent work of Wu et al. (2020), which obtained impressive 3D reconstruction networks with unsupervised learning. However, their algorithm is only applicable to symmetric objects. In this paper, we eliminate the symmetry requirement with a novel unsupervised algorithm that can learn a 3D reconstruction network from a multi-image dataset. Our algorithm is more general and covers the symmetry-required scenario as a special case. Besides, we employ a novel albedo loss that  improves the reconstructed details and realisticity. Our method surpasses the previous work in both quality and robustness, as shown in experiments on datasets of various structures, including single-view, multi-view, image-collection, and video sets. Code is available at: \url{https://github.com/VinAIResearch/LeMul}. 


\end{abstract}

\section{Introduction}

Images are 2D projections of real-world 3D objects, and recovering the 3D structure from a 2D image is an important computer vision task with many applications. 
Most image-based 3D modeling methods rely on multi-view inputs \cite{snavely2006photo,snavely2008modeling,furukawa2010towards,furukawa2007accurate,faugeras2001geometry,zhou2017unsupervised,ummenhofer2017demon,godard2017unsupervised}, requiring multiple images of the target object captured from different views. 
However, these methods are not applicable to the scenarios where only a single input image is available, which is the focus of our work in this paper. This problem is called single-view 3D reconstruction, and it is ill-posed since an image can be a projection of infinitely many 3D shapes. Interestingly, humans are very good at estimating the 3D structure of any known class object from a single image; we can even predict how it looks in unseen views. This is perhaps because humans have strong prior knowledge about the 3D shape and texture of the object class in consideration. Inspired by this observation, many category-specific 3D modeling methods have been proposed for specific object categories such as faces \cite{blanz1999morphable,romdhani2003efficient,Zhu2016Face,tran16_3dmm_cnn,richardson20163d,Tewari_2017_ICCV,tran2018nonlinear,feng2018prn}, hands \cite{zimmermann2017learning,mueller2018ganerated,boukhayma20193d,ge20193d}, and bodies \cite{pavlakos2019expressive,jiang2020coherent}.


In this paper, instead of focusing on any individual category, we aim to develop a general framework that can work for any object category, as long as there are many images from that category to train a single-view 3D reconstruction network. Furthermore, given the difficulty of acquiring 3D ground-truth annotation, we also aim to develop an unsupervised learning method which does not require the ground-truth 3D structures for the objects in the training images. However, this is a challenging problem due to the huge variation of the training images, regarding their viewpoint, appearance, illumination, and background.

A recent study \cite{wu2020unsupervised} made a break-though in solving this problem with a novel end-to-end trainable deep network. Their network consisted of several modules to regress the image formation's components, including the object's 3D shape, texture, viewpoint, and lighting parameters, so that the rendered image was similar to the input. The modules were trained in an unsupervised manner on image datasets. They assumed a single image per training example, so it was still highly under-constrained. To make this training procedure converge, the authors proposed using the symmetry constraint. Their system successfully recovered 3D shape of human faces, cat faces, and synthetic cars after training on respective datasets. For convenience, from now on we will call this \underline{Le}arning from \underline{Sym}metry method as \textbf{\textit{LeSym}}.


While showing good initial results, LeSym has several limitations. First, it requires the target object to be almost symmetric, severely restricting its applicability to certain object classes. For highly asymmetric objects, this method does not work, and for nearly symmetric objects, it would not preserve the asymmetric details. Second, with a strong symmetry constraint, an incorrect mirror line estimation would lead to unrealistic 3D reconstruction. Some examples and detailed discussions on these issues can be found in \Sref{sec:Experiments}. Third, when multiple images of the same object in the training dataset are available, LeSym cannot correlate and leverage these images to improve the reconstruction accuracy and stability. This is a  drawback because there are many imagery datasets that contain multiple images for each object. For example, multiview stereo datasets have photos of each object captured at different views. Some datasets instead have multiple pictures of the same view but with different lighting conditions or focal lengths. Facial datasets often have multiple images for each person, and video datasets have a large number of frames covering the same object in each video. 

In this paper, we propose a more general framework, called \textbf{\textit{LeMul}}, that effectively \underline{Le}arns from \underline{Mul}ti-image datasets for more flexible and reliable unsupervised training of 3D reconstruction networks. It employs loose shape and texture consistency losses based on component swapping across views. \anh{This is an ``unsupervised'' method since it does not require any 3D ground-truth data in training. Although it exploits multiple images per training instance, these images are so diverse and cannot be combined in traditional approaches to form any 3D supervision.} LeMul can cover the symmetric object addressed in LeSym by using the original and the flipped image with less regularized results. More importantly, it handles a wider range of training datasets and object classes. 

Besides, we employ an albedo loss in LeMul, which accurately recovers fine details of the 3D shape. This loss is inspired by a well-known Shape-from-Shading (SfS) literature \cite{or2015rgbd}. It greatly improves the realisticity of the reconstructed 3D model, sometimes approaching laser-scan quality, from a low-res single image input.

In short, our contributions are: (1) we introduce a general framework, called LeMul, that can exploit multi-image datasets in learning 3D object reconstruction from a single image without the symmetry constraint; (2) we employ shape and texture consistency losses to make that unsupervised learning converge; (3) we apply an albedo loss to improve realisticity of the reconstruction results; (4) LeMul shows state-of-the-art performance, qualitatively and quantitively, on a wide range of datasets.


\section{Related Work}
In this section, we briefly review the existing image-based 3D reconstruction approaches, from classical to deep-learning-based algorithms.

\minisection{Multi-view 3D reconstruction}. This approach requires multiple images of the target object captured at different viewpoints. It consists of two sub-tasks: Structure-from-Motion (SfM) and Multi-view Stereo (MVS). SfM estimates from the input images the camera matrices and a sparse 3D reconstructed point-cloud~\cite{snavely2006photo,snavely2008modeling}. SfM requires robust keypoints extracted from each input view for matching and reconstruction. MVS assumes known camera matrices for a dense 3D reconstruction~\cite{furukawa2007accurate,furukawa2010towards}. These tasks are often combined to form end-to-end systems: SfM provides camera matrix estimation as an input to MVS~\cite{wu2011visualsfm}. These approaches were well-studied in classical literature, and they have been further improved with deep learning~\cite{zhou2017unsupervised,ummenhofer2017demon,godard2017unsupervised}. These methods, however, are unfit for our objective of 3D reconstruction from a single image at inference time. Even at training time, they hardly work with our in-the-wild inputs with low image quality, diverse capturing conditions, and freely non-rigid deformation.

\minisection{Shape from X} is another common 3D modeling approach that relies on a specific aspect of the image(s) such as silhouettes \cite{koenderink1984does}, focus \cite{SfF2005}, symmetry \cite{mukherjee1995shape,franccois2003mirror,thrun2005shape,sinha2012detecting}, and shading \cite{zhang1999shape,kemelmacher20103d,barron2014shape,or2015rgbd}. These methods only work on restricted conditions, thus do not apply to in-the-wild data. We focus on two latter directions since they are applicable to our problem. Shape-from-symmetry assumes the target object is symmetric, thus using the original and flipped image as a stereo pair for 3D reconstruction. Shape-from-shading (SfS) relies on some shading model, normally Phong shading \cite{phong1975illumination} or Spherical Harmonic Lighting \cite{green2003spherical}, and solves an inverse rendering problem to decompose image's intrinsic components, including 3D shape, albedo, and illumination. SfS methods often either refine an initial 3D~\cite{kemelmacher20103d,or2015rgbd} or solve an optimization problem with multiple heuristic constraints \cite{barron2014shape}. We are particularly interested in \cite{or2015rgbd}, which employs bilateral-like loss functions to obtain fine-details on an initial raw depth-map. 

\minisection{Deep-learning-based 3D modeling}. Deep learning provides a powerful tool to handle challenging computer vision problems, including 3D reconstruction from a single image. Some studies managed to solve the monocular depth estimation \cite{eigen2015predicting,xu2018structured,ricci2018monocular,fu2018deep} from a single image via supervised learning on ground-truth datasets. Some other studies learned a 3D shape representation from 3D datasets, using a generative model such as GAN or VAE, and fit it into the input image either with or without supervision \cite{choy20163d,girdhar2016learning,zhu2017rethinking,zhu2017rethinking,kundu20183d}. These methods, however, require ground-truth data for supervision or 3D shape datasets for prior learning. They are not unsupervised and cannot handle a new object class that has no available 3D data.

\begin{figure*}[t]
\centering
\includegraphics[width=1.\textwidth]{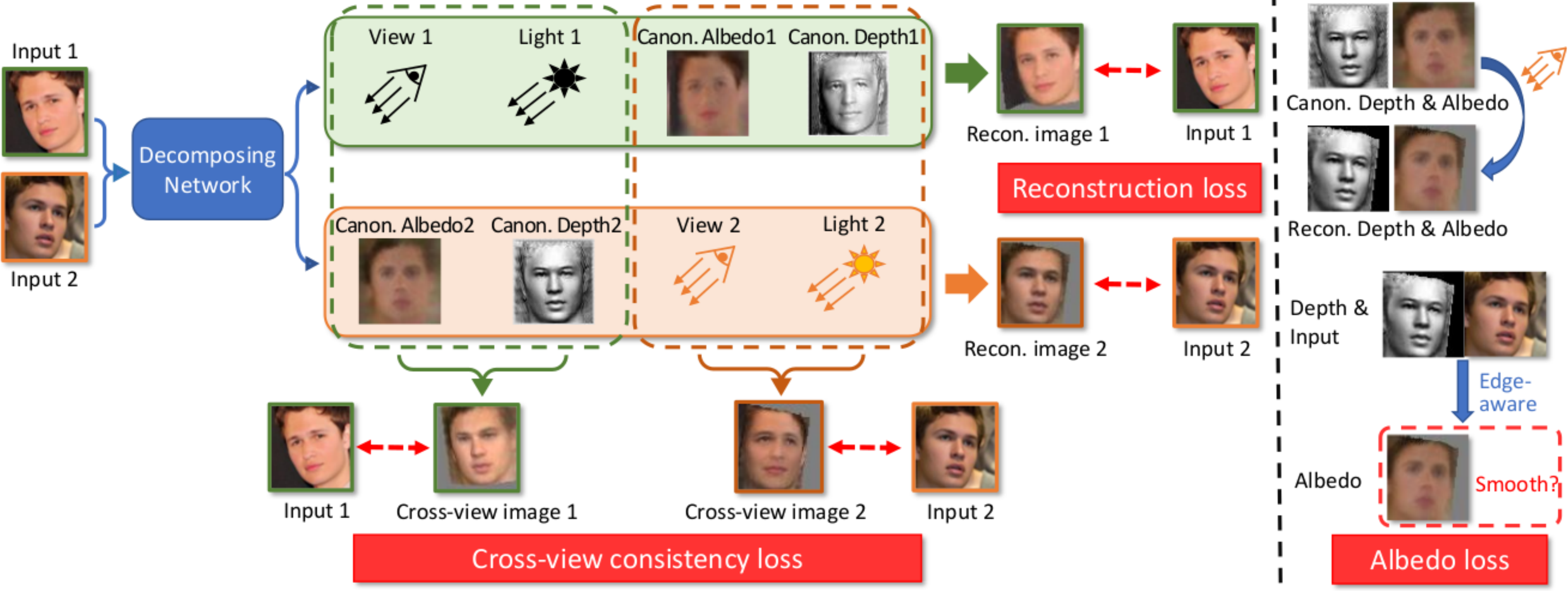}
\vspace{-5mm}
\caption{\textbf{Overview of the proposed system.} \anh{We train a decomposing network to optimize different loss components. Note that we omit the confidence maps in this figure for simplicity. Also, we use diffuse shading images to visualize depth maps.}} 
\vspace{-2mm}
\label{fig:overview}
\end{figure*}

\minisection{Category-specific 3D reconstruction}. Some research focus on reconstructing 3D models of a specific object class, such as human faces \cite{blanz1999morphable,romdhani2003efficient,Zhu2016Face,tran16_3dmm_cnn,richardson20163d,Tewari_2017_ICCV,tran2018nonlinear,feng2018prn}, hands \cite{zimmermann2017learning,mueller2018ganerated,boukhayma20193d,ge20193d}, and bodies \cite{pavlakos2019expressive,jiang2020coherent}. The 3D modeling process often heavily relies on well-defined shape priors. For instance, early 3D face modeling studies used simple PCA models learned from facial landmarks such as Active Appearance Model (AAM) \cite{cootes1998active,cootes2001active} and Constrained Local Model (CLM) \cite{cristinacce2008automatic,baltruvsaitis20123d}. Later, statistical models for 3D face shape and albedo learned from 3D face scans, called 3D Morphable Models (3DMMs) \cite{paysan20093d,gerig2018morphable}, were used as an effective prior in 3D face modeling algorithms \cite{blanz1999morphable,romdhani2003efficient,Zhu2016Face,tran16_3dmm_cnn,richardson20163d,Tewari_2017_ICCV}. Recently, many works have explored other 3D face presentations, such as non-linear 3DMMs \cite{tran2018nonlinear} or GCN-based features \cite{ranjan2018generating,wei20193d}. Instead of learning specific models based on characteristics of each object class, we target a general framework that can extract 3D shape prior for any class just from in-the-wild images.

\minisection{LeSym} \cite{wu2020unsupervised} was the first work that could handle the task of 3D modeling from a single image in a general and unsupervised manner. It followed the SfS approach to extract the image's intrinsic components, including 3D shape, texture, view, and illumination parameters. The network was trained to minimize the reconstruction loss, comparing the rendered image and the input, using a differentiable renderer on a large image set of same-class objects. The optimization problem was under-constrained, so the authors assumed symmetry on the target object and incorporated the flipped image as in Shape-from-Symmetry algorithms. LeSym showed impressive reconstruction results on human faces, cat faces, and synthetic cars. However, the symmetry assumption strongly regularized the estimated 3D models and restricted LeSym's applications. Also, the reconstructed 3D models are still raw, with many  details missing.

\section{Learning from Multi-Image Datasets}
\subsection{Overview}
We revise the mechanism used in LeSym to get LeMul as a more general, effective, and accurate unsupervised 3D reconstruction method. Two key ideas in our proposal are: a multi-image based unsupervised training and a novel albedo loss. The system overview is illustrated in \Fref{fig:overview}.

Unlike LeSym, we do not require the modeling target to be symmetric. Instead, we assume more than one image for each object in the training data. We run the network modules over each image and enforce shape and albedo consistency. Note that having a single image of a symmetric object is a special case of ours; we can simply use the original and flipped input as two images of each training instance, and the 3D model consistency will enforce the object's symmetry. Moreover, this configuration can account for many other common scenarios such as multi-view, multi-exposure, multi-frame datasets. The multi-image configuration is only needed in training. During inference, the system can output a 3D model from a single input image.


Consider a training example and let $\{\I_i\}$ denote the set of~$M$ images of an object taken at different conditions. Each image $\I_i \in R^{H \times W \times 3}$ can be decomposed into four components $(\hat{d}_i,\hat{a}_i,\hat{l}_i,\hat{v}_i)$. The first two components represent the object's 3D model in a \textbf{canonical} view that is independent to camera pose, with $\hat{d}_i \in \mathbb{R}^{H \times W}$ is the depth-map and $\hat{a}_i \in \mathbb{R}^{H \times W \times 3}$ is the albedo-map. The latter components model the capturing conditions, with $\hat{l}_i \in \mathbb{R}^L$ is a vector of $L$ illumination parameters and $\hat{v}_i \in \mathbb{R}^6$ is the viewing vector. The image is formed by a shading function $\mathcal{R}$:
\begin{equation}
\I_i = \mathcal{R} (\hat{d}_i,\hat{a}_i,\hat{l}_i,\hat{v}_i) + \eta_i.
\end{equation}
where $\eta_i$ is the noise term for factors such as background clutter and occlusions. 
The shading model $\mathcal{R}$ is a differentiable renderer \cite{kato2018renderer}, which uses a perspective projection camera, Phong shading model, and Lambertian surface assumption. There are $L{=}4$ illumination parameters, including the weighting coefficients for the ambient term $k_s$ and the diffuse term $k_d$ and the light direction $(l_x, l_y)$. Other details are described in \cite{wu2020unsupervised}.

Our \nhat{\textbf{decomposing network}} consists of four modules to estimate the four intrinsic components $(\hat{d}_i,\hat{a}_i,\hat{l}_i,\hat{v}_i)$ of an input image $I_i$. We denote these modules as $\mathcal{F}_d, \mathcal{F}_a, \mathcal{F}_l$, and $\mathcal{F}_v$ respectively. $\mathcal{F}_d$ and $\mathcal{F}_a$ translate the input to output maps that have the same spatial resolution. $\mathcal{F}_l$ and $\mathcal{F}_v$ are regression networks that output parameter vectors. The outputs of these modules components, denoted as $(d_i,a_i,l_i,v_i)$, are used to reconstruct the input image:
\begin{equation*}
\I_i^r = \mathcal{R} (\mathcal{F}_d(\I_i), \mathcal{F}_a(\I_i), \mathcal{F}_l(\I_i), \mathcal{F}_v(\I_i)) = \mathcal{R} (d_i,a_i,l_i,v_i).
\end{equation*}

There are two desired criteria: (1) the reconstructed image $\I_i^r$ should be similar to the input $\I_i$; (2) for any pair of images coming from the same training sample $\I_i$ and $\I_j$, the estimated canonical depth and albedo maps ($d_i$, $a_i$) and ($d_j$, $a_j$) should be almost similar and interchangeable. These criteria can be formulated into two losses $\mathcal{L}^{rec}$ and $\mathcal{L}^{rec}_{cross}$ respectively. Furthermore, we employ novel loss functions, called albedo losses, inspired by \cite{or2015rgbd} to further improve the reconstruction of fine details. These losses follow the same-view and cross-view settings, and we denote them as $\mathcal{L}^{al}$ and $\mathcal{L}_{cross}^{al}$. The total training loss, therefore, will be:
\begin{equation}
\mathcal{L} = \mathcal{L}^{rec} + \lambda^{cross}\mL^{rec}_{cross} + \lambda^{al}(\mL^{al} + \lambda^{cross}  \mathcal{L}_{cross}^{al}),
\end{equation}
with $\lambda^{al}$ and $\lambda^{cross}$ being weighting hyper-parameters. We will now discuss each loss component above.

\subsection{Reconstruction loss}
We inherit this loss from LeSym. It enforces the reconstructed image to be similar to the input. To discard the effect of the noise $\eta$, another sub-network, called $\mathcal{F}_c$ is used to regress a pair of confidence maps ($c^{l_1}, c^{pe}$) that weigh the pixels in computing the reconstruction loss. The total reconstruction loss is summed over all input views:
\begin{equation*}
\mathcal{L}^{rec}  = \sum_{i=1}^M \left( \mathbb{L}^{l_1}(\I_i, \I_i^r, c^{l_1}_i) + \lambda^{pe} \mathbb{L}^{pe}(g(\I_i), g(\I_i^r), c^{pe}_i)\right),
\end{equation*}
\anh{where $\mathbb{L}^{l_1}$ and $\mathbb{L}^{pe}$ are functions to compute the $l_1$ and perceptual loss components}, ~$g$ is a function to extract the $k$-th layer feature $\mathbf{f}$ of a VGG-16 network pre-trained on ImageNet, and $\lambda^{pe}$ is a weighting hyper-parameter. 

Assuming Gaussian distributions, the mentioned loss components have detailed expressions as following:
\begin{equation}
\mathbb{L}^{l_1}(\I, \I', c)  = \frac{1}{|\Omega|}\sum_{p\in \Omega}\frac{\sqrt{2} |\I(p) - \I'(p)|_1}{c(p)} +  \ln(c(p)),
\end{equation}
\begin{equation}
\mathbb{L}^{pe}(\mathbf{f}, \mathbf{f}', c)  = \frac{1}{|\Omega_k|}\sum_{p\in \Omega_k} \frac{\|\mathbf{f}(p) - \mathbf{f'}(p)\|^2}{2(c(p))^2} + \ln (c(p)),
\end{equation}
with $\Omega$ and $\Omega_k$ as pixel sets in image and feature space.  

\subsection{Cross-view consistency loss}
Reconstruction loss alone is not enough to constrain the reconstruction outcome. Since we have multiple images per training instance, we can enforce the reconstructed 3D models ($d, a$) to be consistent via a cross-view consistency loss. 

In theory, we can simply minimize the distance $\sum_{i \ne j} (\|d_i - d_j\| + \|a_i - a_j\|)$, but we found it ineffective in practice, making the training unstable to converge. Instead, we propose to implement the consistency loss based on a component swapping mechanism. For each pair of views $i \ne j$, we can swap the estimated 3D model from one view ($d_j, a_j$) to the other to render a cross-model image:
\begin{equation}
\I_{ij}^r = \mathcal{R} (d_j,a_j,l_i,v_i).
\end{equation}
This image should be almost the same as the input $\I_i$. Similar to the reconstruction loss, we employ some confidence maps for loss computation. However, these maps correlate two input images ($\I_i, \I_j$), requiring another confidence network. We call this network $\mathcal{F}_{cc}$ that inputs the image pair ($\I_i, \I_j$) stacked by channels and returns a pair of confidence maps ($c_{ij}^{l_1}, c_{ij}^{pe}$). The cross-view loss item for this image pair can be computed as follows:
\begin{equation}
\mathcal{L}_{cross}^{rec}(i, j) = \mathbb{L}^{l_1}(\I_i, \I_{ij}^r, c_{ij}^{l_1}) + \lambda^{pe} \mathbb{L}^{pe}(g(\I_i), g(\I_{ij}^r), c_{ij}^{pe}). \nonumber
\end{equation}

We can compute the cross-view entropy loss for all pairs of $i \ne j$, but this can be computationally expensive if $M$ is large. For computational efficiency, we select the first view as a pivot and use only the pairs related to the first view: 
\begin{equation}
\mathcal{L}^{rec}_{cross} = \sum_{i=2}^M(\mL_{cross}^{rec}(i, 1) + \mL_{cross}^{rec}(1, i)).
\end{equation}

\subsection{Albedo losses}

Although the 3D reconstructed shapes obtained with the above losses are reasonably accurate already, the 3D shapes tend to be over-smooth with many fine details of the 3D surface being inaccurately transferred to the albedo map. For sharper 3D reconstruction, we apply a regularization on the albedo map to avoid overfitting to pixel intensities. This regularization should guarantee that the albedo is smooth at non-edge pixels while preserving the edges. Following \cite{or2015rgbd}, we implement such regularization by albedo loss terms.

An albedo loss requires three aligned inputs, including an input image $\I$ and the corresponding maps for depth $d$ and albedo $a$. It enforces smoothness on $a$:
\begin{equation*}
\mathbb{L}^{al}(\I,a, d) = \frac{1}{|\Omega|} \sum_{p \in \Omega}\big\| {\sum_{p_k \in \mathcal{N}(p)}}{w_k^c w^d_k(a(p) - a(p_k))\big\|^2}. 
\end{equation*}
where $\mathcal{N}(p)$ defines the neighbors of a pixel $p$, $w^c_k$ is the intensity weighting term:
\begin{equation}
w^c_k = \exp\left(-\frac{\|I(p) - I(p_k)\|^2}{2\sigma_c^2}\right),
\end{equation}
and $w^d_k$ is the depth weighting term:
\begin{equation}
w^d_k = \exp\left(-\frac{\|d(p) - d(p_k)\|^2}{2\sigma_d^2}\right).
\end{equation}
The weighting terms suppress the effect of neighbor pixels that likely come from other regions due to a large gap in intensity/depth compared with the current one. We use $\sigma_c$ and $\sigma_d$ to control the allowed intensity and depth discontinuity.

Note that the three inputs of the albedo loss needed to be aligned pixel-by-pixel. We keep the original input $\I$, which is at an estimated view $v$. Therefore, we cannot use the canonical maps ($d, a$) directly, so we transform them to the view $v$. This process can be done by a warping function $\mathcal{W}$. This function first computes the 3D shape from the canonical depth $d$, then project and render it at the view $v$. The outputs are transformed depth and albedo maps:
\begin{equation}
(d^v, a^v) = \mathcal{W}((d,a), d, v).
\end{equation}

Similar to the previous loss terms, we compute the albedo loss in same-view and cross-view settings:
\begin{align}
&\mathcal{L}^{al}  = \sum_{i=1}^{M}{ \mathbb{L}^{al}(\I_i, a_i^{v_i},  d_i^{v_i})},\\
&\mathcal{L}^{al}_{cross} = \sum_{i=2}^M \left( \mathbb{L}^{al}(\I_1, a_i^{v_1}, d_i^{v_1}) + \mathbb{L}^{al}(\I_i, a_1^{v_i}, d_1^{v_i})\right). \nonumber 
\end{align}


\section{Experiments}\label{sec:Experiments}
\subsection{Experimental setups}
\subsubsection{Implementation details}
We implemented our system in PyTorch. The networks $\mathcal{F}_d, \mathcal{F}_a, \mathcal{F}_l, \mathcal{F}_v$, and $\mathcal{F}_c$ had the same structure as in the official released code of LeSym\footnote{\url{https://github.com/elliottwu/unsup3d}}. The cross-view confidence network $\mathcal{F}_{cc}$ was similar to $\mathcal{F}_{c}$, except for having six input channels instead of three. In all experiments, we used the same input image size $H {=} W {=} 64$. The hyper-parameters were set as $\lambda^{pe} {=} 1$, $\lambda^{cross} {=} \lambda^{al} {=} 0.5$, $\sigma_c {=} 0.05$ and $\sigma_d {=} 2$.  The networks were jointly trained with Adam optimizer at a fixed learning rate $0.0001$ until convergence.

\subsubsection{Datasets}
To evaluate the proposed algorithm, we run experiments on datasets with various capturing settings and data structures (single-view, multi-view, image-collection, or video):

\minisection{BFM} is a synthetic dataset of 200K human face images proposed by LeSym. Each image is rendered with a 3D shape and texture randomly sampled from the Basel Face Model \cite{paysan20093d}, a random view, and one of the spherical harmonics lights estimated from CelebA images \cite{liu2015deep}. Besides RGB images, the ground-truth 3D depth-maps are also provided. We use this dataset to quantitatively evaluate our approach as well as comparing it with other baselines.


\minisection{CelebA \cite{liu2015deep}} is a popular facial dataset of more than 200K celebrity images. The images were captured under in-the-wild conditions. It is split into three subsets for  training, validation, and testing with 162K, 20K, and 20K images, respectively. We use this dataset to compare LeMul and LeSym under the ``single-view'' and ``symmetric-objects'' settings. We generate two image inputs for each training instance, including the original and the flipped image.

\minisection{Cat Faces} is a dataset of 11.2K images capturing cat faces in-the-wild. This dataset was constructed in LeSym by combining two previous datasets \cite{zhang2008cat,parkhi2012cats}. This set is split into 8930 training and 2256 testing instances. This dataset is also under the ``single-view'' and ``symmetric-objects'' settings, and its two-view data is formed similar to CelebA.

\minisection{Multi-PIE \cite{multiPIE}} is a large human-face dataset captured in studio settings. It contains more than 750K images of 337 people involved in from one to four different recording sessions. In each session, each subject has a collection of images captured at 15 view-points, 19 illuminations, and with several expressions. We excluded images with extreme light or overwhelmed expression and selected ones at three viewing angles corresponding to frontal, $15^{\circ}$-to-the-left, and $15^{\circ}$-to-the-right views, to form a multi-view image set. Each training instance is a set of three images of each person, captured at the selected views. We use random illumination, causing three input views drastically different and unable to be used by traditional multi-view stereo methods. 

\minisection{CASIA-WebFace \cite{yi2014learning}} has 500K face images of 10K people collected from the Internet. Each person has on average 50 in-the-wild images with drastically different conditions. We keep the last 200 subjects for testing and use the rest for training. In each training epoch, for each subject, we randomly select $M {=} 3$ images of that person regardless of pose, expression, and illumination to form a training example.

\minisection{YouTube Faces (YTF) \cite{wolf2011face}} is a video dataset that consists of 3425 videos of 1595 people. The videos have low-quality frames, which were severely degraded by video compression. Many videos are also bad for 3D face modeling, with the target faces at non-frontal views and barely moving. Still, we aim to evaluate our method on such extreme conditions. For each video, we extract the frames and crop them around the target faces. We split the videos for training~(3299) and testing (126). Similar to CASIA, in each training epoch and with each video, we randomly select $M {=} 3$ frames to form a training instance.

\minisection{Quantitative Metrics.}
 \anh{For fair comparison results, we use the same metrics used in LeSym. The first metric is \textit{Scale-Invariant Depth Error} (SIDE) \cite{NIPS2014_7bccfde7}, which computes the standard deviation of the difference between the estimated depth map at the input view and the ground-truth depth map at the log scale.} 
However, we argue that this metric is not a strong indicator of the reconstruction quality. A reasonable error in the object distance estimation, while not affecting the projected image, can cause SIDE varying. In contrast, it is ineffective in evaluating the reconstructed surface quality. We can smooth out the depth-map or add small random noise to it but cause a minimal change in SIDE value.

Instead, we focus on the second metric, which is the \textbf{\textit{Mean Angle Deviation} (MAD) \cite{wu2020unsupervised}} between normal maps computed from estimated depth map $d^v$ and ground-truth depth map $d^*$. It can measure how well the surface is captured and is sensitive to surface noise.

\setlength{\tabcolsep}{4pt}
\begin{table}[!tbp]
    \centering
    \begin{tabular}{c l c c}
        \toprule 
        No & Baseline & SIDE(x$10^{-2}$)$\downarrow$ & MAD(deg.)$\downarrow$ \\ 
        \midrule 
        (1) & Supervised & $0.410\pm0.103$ & $10.78\pm1.01$  \\ \hline
        (2) & Const. null depth & $2.723\pm0.371$ & $43.34\pm2.25$  \\
        (3) & Average G.T. depth & $1.990\pm0.556$ & $23.26\pm2.85$  \\ 
        (4) & LeSym & \textbf{0.793}$\pm$\textbf{0.140} & $16.51\pm1.56$  \\ 
        (5) & LeMul (proposed) & $0.834\pm0.169$ & \textbf{15.49$\pm$1.50}  \\ \bottomrule 
    \end{tabular}
    \vskip -0.1in
    \caption{BFM results comparison with baselines.}
    \label{tab:BFM_result}
    \vspace{-4mm}
\end{table}

\subsection{Quantitative experiments}
In this section, we perform quantitative evaluations on the BFM dataset with provided ground-truth data.

\minisection{BFM results.} We trained and tested our algorithm on the BFM dataset, and the results are reported in Table \ref{tab:BFM_result}, along with some baselines: (1) supervised 3D reconstruction network as the upper bound, (2) a dummy network returning a constant null depth, (3) a dummy network producing a constant mean depth computed over the ground-truth one, and (4) LeSym. As can be seen, LeMul outperforms the dummy networks by a wide margin. Compared with LeSym, it achieves a better MAD number with $1^\circ$ decrease, implying better reconstructed 3D surfaces with details recovered.

\begin{figure}[t]
\centering
\vspace{-2mm}
\subfloat[]{
\includegraphics[width=.2\textwidth]{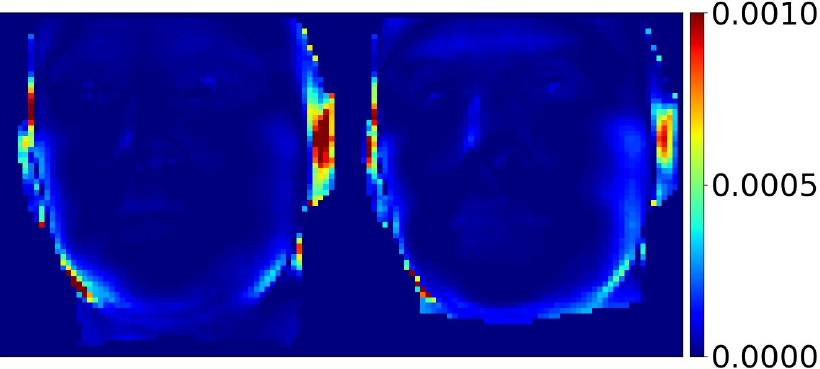}
\vspace{-2mm}
\label{fig:side}
}
\hspace{2mm}
\subfloat[]{
\includegraphics[width=.188\textwidth]{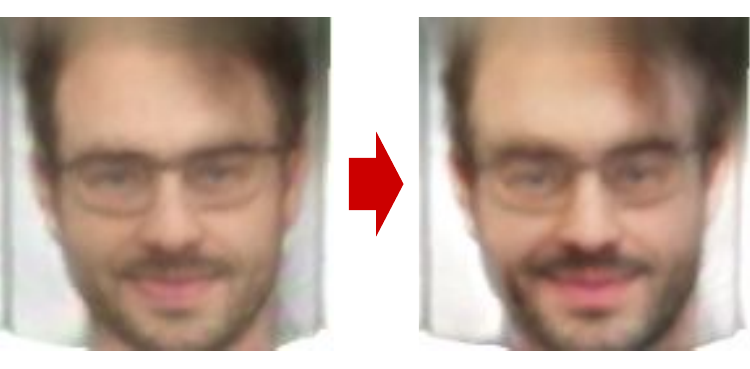}
\vspace{-2mm}
\label{fig:fintuned}
}
\vspace{-3mm}
\caption{\textbf{Qualitative analyses.} (a) LeMul vs. LeSym (SIDE) and (b) Texture refinement.}
\vspace{-2mm}
\label{fig:misc}
\end{figure}

\anh{As for SIDE, we examine the error maps and find that LeMul provides a better overall depth estimation. However, the outliers, particularly on the face boundary or outer components (ears, neck), are more unstable and skew the average score. If we compute the SIDE metric over the facial region bounded by Dlib's 68-landmarks, LeMul has a lower error (\textbf{\textit{0.00534}}) compared with LeSym (\textbf{\textit{0.00564}}), confirming this observation. Fig. \ref{fig:side} provides a common scenario in which LeMul provides a lower error on most facial areas but higher errors on the boundary and an ear.}

\minisection{Ablation studies.} We run ablation experiments to evaluate each proposed component's contribution to our result on the BFM dataset. From LeSym as the baseline, we can modify it to follow our multi-view scheme or integrate the albedo losses. As reported in Table \ref{tab:ablation}, each of our proposals positively affects the MAD numbers. We achieve the best reconstructed 3D surfaces when combining both techniques. 

\begin{table}[!tbp]
    \centering
    \begin{tabular}{c l c c}
        \toprule 
        No & Method & SIDE(x$10^{-2}$)$\downarrow$ & MAD(deg.)$\downarrow$ \\ \midrule 
        (1) & Baseline \cite{wu2020unsupervised} & $0.793\pm0.140$ & $16.51\pm1.56$  \\ 
        (2) & \hspace{1ex} + multi-view & \textbf{0.728}$\pm$\textbf{0.135} & $15.73\pm1.54$  \\
        (3) & \hspace{1ex} + albedo loss & $0.899\pm0.217$ & $16.35\pm1.79$  \\ 
        (4) & \hspace{1ex} + mul+al (full) & $0.834\pm0.169$ & \textbf{15.49$\pm$1.50}  \\ \bottomrule 
    \end{tabular}
    \vskip -0.1in
    \caption{\textbf{Ablation studies} on BFM dataset}
    \label{tab:ablation}
    \vspace{-3mm}
\end{table}

\subsection{Qualitative experiments}
\begin{figure*}[t]
\centering
\includegraphics[width=.9\textwidth]{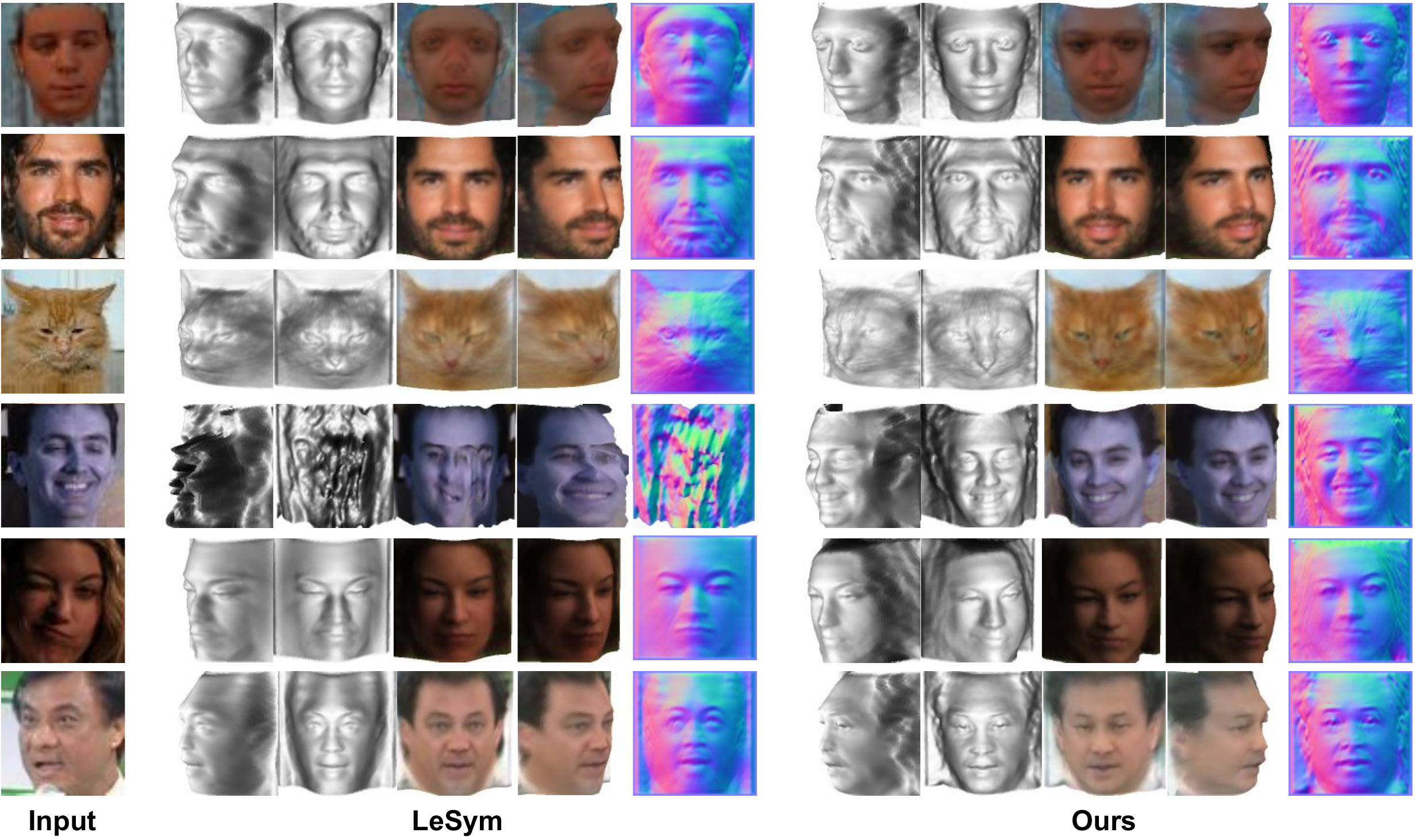}
\vspace{-2mm}
\caption{Comparing the reconstructed 3D models from the baseline method LeSym model LeSym \cite{wu2020unsupervised} and ours. The datasets from top to bottom: BFM \cite{wu2020unsupervised}, CelebA \cite{liu2015deep}, Cat Faces \cite{wu2020unsupervised}, Multi-PIE \cite{multiPIE}, CASIA-WebFace \cite{yi2014learning}, and Youtube Faces \cite{wolf2011face}. For each 3D model, we provide two textureless views, two textured views, and the canonical normal map.}
\label{fig:qual}
\end{figure*}

We qualitatively compare our method to the LeSym baseline  on the mentioned datasets in \Fref{fig:teaser} and \Fref{fig:qual}. In all experiments, LeSym uses each single image as a training instance and applies the symmetry constraint. Our method also assumes that symmetry property on BFM, CelebA, and Cat Faces by using pairs of original and flipped images as training instances. However, on Multi-PIE, CASIA-WebFace, and Youtube Faces, we completely drop that assumption and use multi-image examples in training. 

\minisection{Three symmetry-assumed datasets}. On BFM, CelebA, and Cat Faces, both LeSym and LeMul can reconstruct reasonable 3D models. However, thanks to the albedo loss, LeMul can recover more 3D details such as human hairs, beards, and cat furs. The 3D models are well recognizable even without texture.

\minisection{Multi-PIE results}. LeSym completely collapsed, perhaps due to the limited number of poses and the asymmetric lights. LeMul, instead, performed well on this data configuration with high quality produced 3D models. 

\minisection{CASIA-WebFace results}. LeMul can learn well the 3D face structure. It is impressive since the images used in each training example are wildly different; they are even challenging for humans to correlate, as illustrated in \Fref{fig:teaser} (second row). In both \Fref{fig:teaser} and \Fref{fig:qual}, LeMul can capture asymmetric details such as one-sided hairstyle and lopsided smile. In contrast, LeSym over-regulated the 3D shapes with the symmetry constraint, producing incorrect 3Ds.

\minisection{Youtube Faces results}. This dataset is pretty challenging to our training due to low-quality images and limited variation between frames in each video. Still, LeMul manages to converge and produce reasonable results at test time. When the input image is not too blurry, LeMul can reconstruct a 3D model with more details compared with LeSym, while it does not suffer from the symmetry assumption.

\subsection{User surveys} 
We further compared our method with the baseline via user surveys. We skipped this test on BFM, which was already used in quantitative evaluations, and Multi-PIE, in which LeSym completely failed. For each remaining dataset, we created a survey with $30$ testing images randomly sampled from the respective test set. We generated two 3D models, estimated by LeSym and LeMul, for each image and produced corresponding videos to illustrate these models in various viewing angles. Each tester was asked to pick which model was better. At least 40 people took each survey, leading to at least 1200 answers per dataset.

We report the rate that each method is selected in Table \ref{tab:user}. LeMul outperforms LeSym on CelebA and CASIA datasets by a wide margin, showing that LeMul can recover better 3D models in good training conditions. Notably, it was selected near $80\%$ of time on CASIA, proving the superiority of the multi-image setting over the symmetry constraint. It also beats LeSym on YTF and Cat Faces datasets but with smaller gaps. We found many YTF frames too blurry, making both 3D models smooth and hard to compare. The small gap of LeMul over LeSym came from clear frames, which is still very meaningful. Finally, on Cat Faces, while our models are more detailed, some testers preferred smooth 3Ds from LeSym, decreasing our selected rate. This phenomenon suggests that it is not always good to have many details, opening a future research to improve our method.
\begin{table}[!tbp]
    \centering
    \begin{tabular}{l c c c c}
        \hline
        Method & CelebA & CatFaces & CASIA & YTF \\ \hline
        LeSym \cite{wu2020unsupervised} & 36.01 & 47.03  & 20.86& 45.23\\
        Ours & \textbf{63.99} & \textbf{52.97} & \textbf{79.14} & \textbf{54.77}\\ \hline
    \end{tabular}
    \vskip -0.1in
    \caption{\textbf{User survey results.} For each dataset, we report the rate (\%) that each method is selected by the tester for providing a better 3D model.}
    \label{tab:user}
    \vspace{-4mm}
\end{table}

\begin{figure*}[t]
\centering
\includegraphics[width=.9\textwidth]{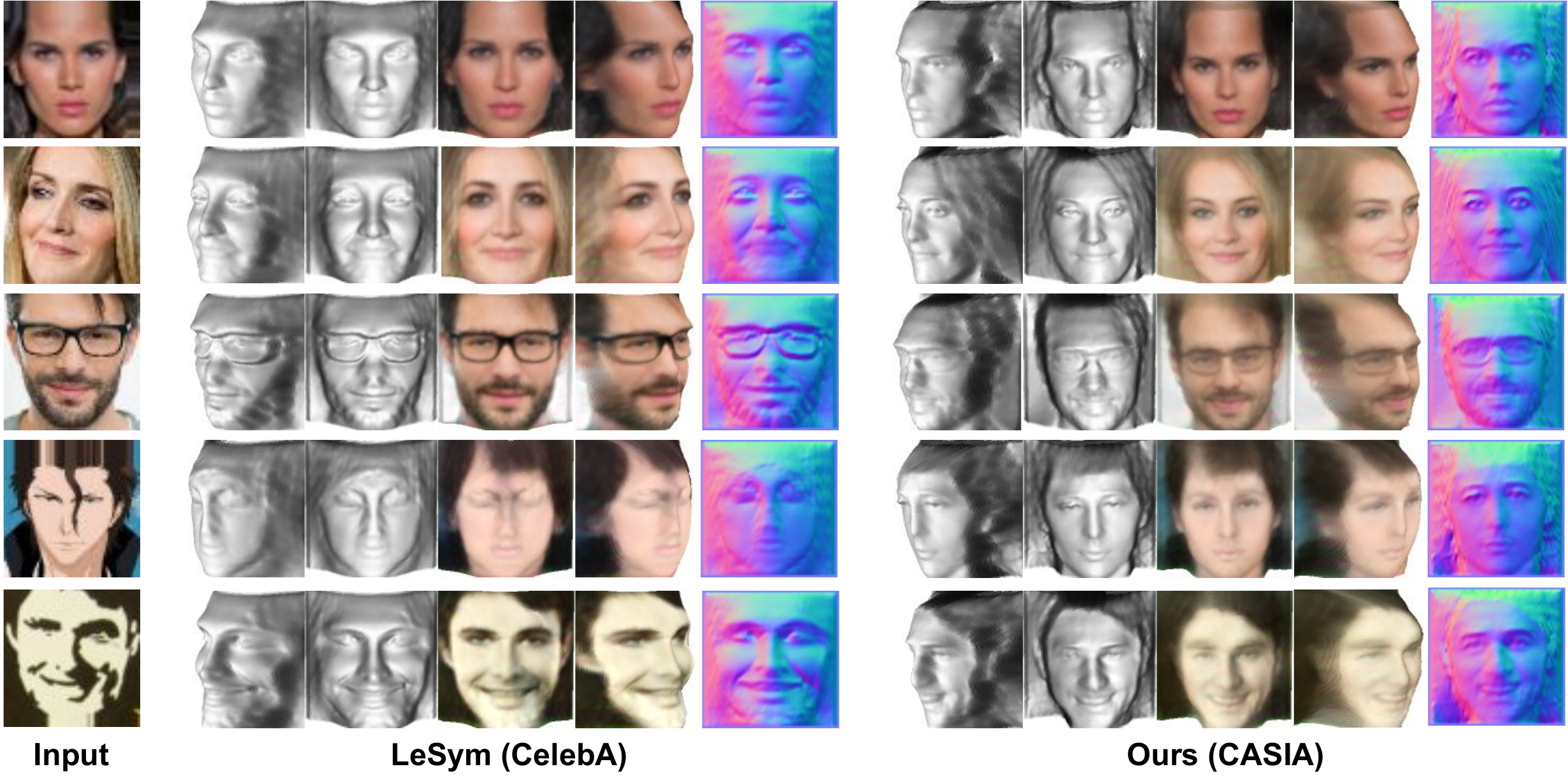}
\vspace{-2mm}
\caption{Reconstructed 3D models from in-the-wild images. We compare the baseline model LeSym \cite{wu2020unsupervised} trained on CelebA dataset \cite{liu2015deep}, and our method trained on CASIA-WebFace \cite{yi2014learning} dataset. For each 3D model, we provide two textureless views, two textured views, and the canonical normal map.}
\label{fig:wild}
\vspace{-2mm}
\end{figure*}

\subsection{In-the-wild tests}
Finally, we run evaluation on in-the-wild facial images collected from the Internet. We select the LeMul model trained on the CASIA dataset since it can capture even the asymmetric details. In contrast, among LeSym models for human face, the released model trained on the CelebA dataset shows the best reconstruction quality. We compare these models on some in-the-wild images in \Fref{fig:wild}. The 3D shapes generated from LeSym are often distorted by symmetry regulation. Our results, instead, look more natural and detailed. Particularly, LeMul can create a realistic-looking 3D model from a cartoon drawing (the fourth row).

\anh{\subsection{Texture refinement.}} 
\anh{We observe that the regressed texture with models trained on CASIA and YTF datasets is a bit blurry, possibly due to two reasons. First, these datasets have lower image quality compared to CelebA; many images have blur, noise, or JPEG artifacts. Second, the models have to learn the subject's albedo from vastly different inputs, causing blurry texture. We propose a simple solution to fix the second issue. After getting a trained model, we can finetune $\mathcal{F}_a$, $\mathcal{F}_l$, and $\mathcal{F}_c$ while freezing the other modules for a few epochs on single-image inputs of the same training set. As shown in Fig. \ref{fig:fintuned}, the estimated texture is significantly improved. Note that this refinement preserves the high-quality 3D shape evaluated in previous experiments.} \\

\vspace{-4mm}
\section{Discussions}
\vspace{-1mm}
In this paper, we present a novel system that shows the state-of-the-art 3D modeling quality in unsupervised learning for single-view 3D object reconstruction. The key insights are to exploit multi-image datasets in training and to employ albedo losses for improved detailed reconstruction. 

\anh{Our method can work on various training datasets ranging from single- and multi-view datasets to image collection and video data. However, a current limitation of our work is that the images of the target object need to be compatible to the depth-map representation, being primarily frontal view without self-occlusion. We plan to address this limitation in future work to increase the applicability of our method.} 


{\small
\bibliographystyle{ieee_fullname}
\bibliography{longstrings, egbib}

\begin{thebibliography}{10}\itemsep=-1pt

\bibitem{baltruvsaitis20123d}
Tadas Baltru{\v{s}}aitis, Peter Robinson, and Louis-Philippe Morency.
\newblock 3d constrained local model for rigid and non-rigid facial tracking.
\newblock In {\em Proceedings of the {IEEE} Conference on Computer Vision and
  Pattern Recognition}, 2012.

\bibitem{barron2014shape}
Jonathan~T Barron and Jitendra Malik.
\newblock Shape, illumination, and reflectance from shading.
\newblock {\em IEEE Transactions on Pattern Analysis and Machine Intelligence},
  pages 1670--1687, 2014.

\bibitem{blanz1999morphable}
V. Blanz and T. Vetter.
\newblock Morphable model for the synthesis of {3D} faces.
\newblock In {\em Proceedings of the ACM SIGGRAPH Conference on Computer
  Graphics}, 1999.

\bibitem{boukhayma20193d}
Adnane Boukhayma, Rodrigo~de Bem, and Philip~HS Torr.
\newblock 3d hand shape and pose from images in the wild.
\newblock In {\em Proceedings of the {IEEE} Conference on Computer Vision and
  Pattern Recognition}, 2019.

\bibitem{choy20163d}
Christopher~B Choy, Danfei Xu, JunYoung Gwak, Kevin Chen, and Silvio Savarese.
\newblock 3d-r2n2: A unified approach for single and multi-view 3d object
  reconstruction.
\newblock In {\em Proceedings of the European Conference on Computer Vision},
  2016.

\bibitem{cootes1998active}
Timothy~F Cootes, Gareth~J Edwards, and Christopher~J Taylor.
\newblock Active appearance models.
\newblock In {\em Proceedings of the European Conference on Computer Vision},
  1998.

\bibitem{cootes2001active}
Timothy~F. Cootes, Gareth~J. Edwards, and Christopher~J. Taylor.
\newblock Active appearance models.
\newblock {\em IEEE Transactions on Pattern Analysis and Machine Intelligence},
  pages 681--685, 2001.

\bibitem{cristinacce2008automatic}
David Cristinacce and Tim Cootes.
\newblock Automatic feature localisation with constrained local models.
\newblock {\em Pattern Recognition}, pages 3054--3067, 2008.

\bibitem{eigen2015predicting}
David Eigen and Rob Fergus.
\newblock Predicting depth, surface normals and semantic labels with a common
  multi-scale convolutional architecture.
\newblock In {\em Proceedings of the International Conference on Computer
  Vision}, 2015.

\bibitem{NIPS2014_7bccfde7}
David Eigen, Christian Puhrsch, and Rob Fergus.
\newblock Depth map prediction from a single image using a multi-scale deep
  network.
\newblock In Z. Ghahramani, M. Welling, C. Cortes, N. Lawrence, and K.~Q.
  Weinberger, editors, {\em Advances in Neural Information Processing Systems}.
  Curran Associates, Inc., 2014.

\bibitem{faugeras2001geometry}
Olivier Faugeras, Quang-Tuan Luong, and Theo Papadopoulo.
\newblock {\em The geometry of multiple images: the laws that govern the
  formation of multiple images of a scene and some of their applications}.
\newblock 2001.

\bibitem{SfF2005}
P. {Favaro} and S. {Soatto}.
\newblock A geometric approach to shape from defocus.
\newblock {\em IEEE Transactions on Pattern Analysis and Machine Intelligence},
  pages 406--417, 2005.

\bibitem{feng2018prn}
Yao Feng, Fan Wu, Xiaohu Shao, Yanfeng Wang, and Xi Zhou.
\newblock Joint 3d face reconstruction and dense alignment with position map
  regression network.
\newblock In {\em Proceedings of the European Conference on Computer Vision},
  2018.

\bibitem{franccois2003mirror}
Alexandre~RJ Fran{\c{c}}ois, G{\'e}rard~G Medioni, and Roman Waupotitsch.
\newblock Mirror symmetry 2-view stereo geometry.
\newblock {\em Image and Vision Computing}, pages 137--143, 2003.

\bibitem{fu2018deep}
Huan Fu, Mingming Gong, Chaohui Wang, Kayhan Batmanghelich, and Dacheng Tao.
\newblock Deep ordinal regression network for monocular depth estimation.
\newblock In {\em Proceedings of the {IEEE} Conference on Computer Vision and
  Pattern Recognition}, 2018.

\bibitem{furukawa2010towards}
Yasutaka Furukawa, Brian Curless, Steven~M Seitz, and Richard Szeliski.
\newblock Towards internet-scale multi-view stereo.
\newblock In {\em Proceedings of the {IEEE} Conference on Computer Vision and
  Pattern Recognition}, 2010.

\bibitem{furukawa2007accurate}
Yasutaka Furukawa and Jean Ponce.
\newblock Accurate, dense, and robust multi-view stereopsis (pmvs).
\newblock In {\em Proceedings of the {IEEE} Conference on Computer Vision and
  Pattern Recognition}, 2007.

\bibitem{ge20193d}
Liuhao Ge, Zhou Ren, Yuncheng Li, Zehao Xue, Yingying Wang, Jianfei Cai, and
  Junsong Yuan.
\newblock 3d hand shape and pose estimation from a single rgb image.
\newblock In {\em Proceedings of the {IEEE} Conference on Computer Vision and
  Pattern Recognition}, 2019.

\bibitem{gerig2018morphable}
Thomas Gerig, Andreas Morel-Forster, Clemens Blumer, Bernhard Egger, Marcel
  Luthi, Sandro Sch{\"o}nborn, and Thomas Vetter.
\newblock Morphable face models-an open framework.
\newblock In {\em Proceedings of the International Conference on Automatic Face
  and Gesture Recognition}, 2018.

\bibitem{girdhar2016learning}
Rohit Girdhar, David~F Fouhey, Mikel Rodriguez, and Abhinav Gupta.
\newblock Learning a predictable and generative vector representation for
  objects.
\newblock In {\em Proceedings of the European Conference on Computer Vision},
  2016.

\bibitem{godard2017unsupervised}
Cl{\'e}ment Godard, Oisin Mac~Aodha, and Gabriel~J Brostow.
\newblock Unsupervised monocular depth estimation with left-right consistency.
\newblock In {\em Proceedings of the {IEEE} Conference on Computer Vision and
  Pattern Recognition}, 2017.

\bibitem{green2003spherical}
Robin Green.
\newblock Spherical harmonic lighting: The gritty details.
\newblock 2003.

\bibitem{multiPIE}
R. Gross, I. Matthews, J. Cohn, T. Kanade, and S. Baker.
\newblock Multi-pie.
\newblock In {\em Proceedings of the International Conference on Automatic Face
  and Gesture Recognition}, 2008.

\bibitem{jiang2020coherent}
Wen Jiang, Nikos Kolotouros, Georgios Pavlakos, Xiaowei Zhou, and Kostas
  Daniilidis.
\newblock Coherent reconstruction of multiple humans from a single image.
\newblock In {\em Proceedings of the {IEEE} Conference on Computer Vision and
  Pattern Recognition}, 2020.

\bibitem{kato2018renderer}
Hiroharu Kato, Yoshitaka Ushiku, and Tatsuya Harada.
\newblock Neural 3d mesh renderer.
\newblock In {\em Proceedings of the {IEEE} Conference on Computer Vision and
  Pattern Recognition}.

\bibitem{kemelmacher20103d}
Ira Kemelmacher-Shlizerman and Ronen Basri.
\newblock 3d face reconstruction from a single image using a single reference
  face shape.
\newblock {\em IEEE Transactions on Pattern Analysis and Machine Intelligence},
  pages 394--405, 2010.

\bibitem{koenderink1984does}
Jan~J Koenderink.
\newblock What does the occluding contour tell us about solid shape?
\newblock {\em Perception}, pages 321--330, 1984.

\bibitem{kundu20183d}
Abhijit Kundu, Yin Li, and James~M Rehg.
\newblock 3d-rcnn: Instance-level 3d object reconstruction via
  render-and-compare.
\newblock In {\em Proceedings of the {IEEE} Conference on Computer Vision and
  Pattern Recognition}, 2018.

\bibitem{liu2015deep}
Ziwei Liu, Ping Luo, Xiaogang Wang, and Xiaoou Tang.
\newblock Deep learning face attributes in the wild.
\newblock In {\em Proceedings of the International Conference on Computer
  Vision}, 2015.

\bibitem{mueller2018ganerated}
Franziska Mueller, Florian Bernard, Oleksandr Sotnychenko, Dushyant Mehta,
  Srinath Sridhar, Dan Casas, and Christian Theobalt.
\newblock Ganerated hands for real-time 3d hand tracking from monocular rgb.
\newblock In {\em Proceedings of the {IEEE} Conference on Computer Vision and
  Pattern Recognition}, 2018.

\bibitem{mukherjee1995shape}
Dipti~Prasad Mukherjee, Andrew~Peter Zisserman, Michael Brady, and FT Smith.
\newblock Shape from symmetry: Detecting and exploiting symmetry in affine
  images.
\newblock {\em Philosophical Transactions of the Royal Society of London.
  Series A: Physical and Engineering Sciences}, pages 77--106, 1995.

\bibitem{or2015rgbd}
Roy Or-El, Guy Rosman, Aaron Wetzler, Ron Kimmel, and Alfred~M Bruckstein.
\newblock Rgbd-fusion: Real-time high precision depth recovery.
\newblock In {\em Proceedings of the {IEEE} Conference on Computer Vision and
  Pattern Recognition}, 2015.

\bibitem{parkhi2012cats}
Omkar~M Parkhi, Andrea Vedaldi, Andrew Zisserman, and CV Jawahar.
\newblock Cats and dogs.
\newblock In {\em Proceedings of the {IEEE} Conference on Computer Vision and
  Pattern Recognition}, 2012.

\bibitem{pavlakos2019expressive}
Georgios Pavlakos, Vasileios Choutas, Nima Ghorbani, Timo Bolkart, Ahmed~AA
  Osman, Dimitrios Tzionas, and Michael~J Black.
\newblock Expressive body capture: 3d hands, face, and body from a single
  image.
\newblock In {\em Proceedings of the {IEEE} Conference on Computer Vision and
  Pattern Recognition}, 2019.

\bibitem{paysan20093d}
Pascal Paysan, Reinhard Knothe, Brian Amberg, Sami Romdhani, and Thomas Vetter.
\newblock A 3d face model for pose and illumination invariant face recognition.
\newblock In {\em Proceedings of International Conference on Advanced Video and
  Signal based Surveillance}, 2009.

\bibitem{phong1975illumination}
Bui~Tuong Phong.
\newblock Illumination for computer generated pictures.
\newblock {\em Communications of the ACM}, pages 311--317, 1975.

\bibitem{ranjan2018generating}
Anurag Ranjan, Timo Bolkart, Soubhik Sanyal, and Michael~J Black.
\newblock Generating 3d faces using convolutional mesh autoencoders.
\newblock In {\em Proceedings of the European Conference on Computer Vision},
  2018.

\bibitem{ricci2018monocular}
Elisa Ricci, Wanli Ouyang, Xiaogang Wang, Nicu Sebe, et~al.
\newblock Monocular depth estimation using multi-scale continuous crfs as
  sequential deep networks.
\newblock {\em IEEE Transactions on Pattern Analysis and Machine Intelligence},
  pages 1426--1440, 2018.

\bibitem{richardson20163d}
Elad Richardson, Matan Sela, and Ron Kimmel.
\newblock 3d face reconstruction by learning from synthetic data.
\newblock In {\em Proceedings of International Conference on 3D Vision}, 2016.

\bibitem{romdhani2003efficient}
Sami Romdhani and Thomas Vetter.
\newblock Efficient, robust and accurate fitting of a {3D} morphable model.
\newblock In {\em Proceedings of the International Conference on Computer
  Vision}, 2003.

\bibitem{sinha2012detecting}
Sudipta~N Sinha, Krishnan Ramnath, and Richard Szeliski.
\newblock Detecting and reconstructing 3d mirror symmetric objects.
\newblock In {\em Proceedings of the European Conference on Computer Vision},
  2012.

\bibitem{snavely2006photo}
Noah Snavely, Steven~M Seitz, and Richard Szeliski.
\newblock Photo tourism: exploring photo collections in 3d.
\newblock In {\em Proceedings of the ACM SIGGRAPH Conference on Computer
  Graphics}. 2006.

\bibitem{snavely2008modeling}
Noah Snavely, Steven~M Seitz, and Richard Szeliski.
\newblock Modeling the world from internet photo collections.
\newblock {\em International Journal of Computer Vision}, pages 189--210, 2008.

\bibitem{Tewari_2017_ICCV}
Ayush Tewari, Michael Zollhofer, Hyeongwoo Kim, Pablo Garrido, Florian Bernard,
  Patrick Perez, and Christian Theobalt.
\newblock {MoFA}: Model-based deep convolutional face autoencoder for
  unsupervised monocular reconstruction.
\newblock In {\em Proceedings of the International Conference on Computer
  Vision}, 2017.

\bibitem{thrun2005shape}
Sebastian Thrun and Ben Wegbreit.
\newblock Shape from symmetry.
\newblock In {\em Proceedings of the International Conference on Computer
  Vision}, 2005.

\bibitem{tran16_3dmm_cnn}
Anh Tran, Tal Hassner, Iacopo Masi, and G\'{e}rard Medioni.
\newblock Regressing robust and discriminative {3D} morphable models with a
  very deep neural network.
\newblock In {\em Proceedings of the {IEEE} Conference on Computer Vision and
  Pattern Recognition}, 2017.
\newblock
  Available:~\url{http://www.openu.ac.il/home/hassner/projects/CNN3DMM/}.

\bibitem{tran2018nonlinear}
Luan Tran and Xiaoming Liu.
\newblock Nonlinear 3d face morphable model.
\newblock In {\em Proceedings of the {IEEE} Conference on Computer Vision and
  Pattern Recognition}, 2018.

\bibitem{ummenhofer2017demon}
Benjamin Ummenhofer, Huizhong Zhou, Jonas Uhrig, Nikolaus Mayer, Eddy Ilg,
  Alexey Dosovitskiy, and Thomas Brox.
\newblock Demon: Depth and motion network for learning monocular stereo.
\newblock In {\em Proceedings of the {IEEE} Conference on Computer Vision and
  Pattern Recognition}, 2017.

\bibitem{wei20193d}
Huawei Wei, Shuang Liang, and Yichen Wei.
\newblock 3d dense face alignment via graph convolution networks.
\newblock {\em arXiv preprint arXiv:1904.05562}, 2019.

\bibitem{wolf2011face}
Lior Wolf, Tal Hassner, and Itay Maoz.
\newblock Face recognition in unconstrained videos with matched background
  similarity.
\newblock In {\em Proceedings of the {IEEE} Conference on Computer Vision and
  Pattern Recognition}, 2011.

\bibitem{wu2011visualsfm}
Changchang Wu et~al.
\newblock Visualsfm: A visual structure from motion system.

\bibitem{wu2020unsupervised}
Shangzhe Wu, Christian Rupprecht, and Andrea Vedaldi.
\newblock Unsupervised learning of probably symmetric deformable 3d objects
  from images in the wild.
\newblock In {\em Proceedings of the {IEEE} Conference on Computer Vision and
  Pattern Recognition}, 2020.

\bibitem{xu2018structured}
Dan Xu, Wei Wang, Hao Tang, Hong Liu, Nicu Sebe, and Elisa Ricci.
\newblock Structured attention guided convolutional neural fields for monocular
  depth estimation.
\newblock In {\em Proceedings of the {IEEE} Conference on Computer Vision and
  Pattern Recognition}, 2018.

\bibitem{yi2014learning}
Dong Yi, Zhen Lei, Shengcai Liao, and Stan~Z Li.
\newblock Learning face representation from scratch.
\newblock {\em arXiv preprint arXiv:1411.7923}, 2014.
\newblock
  Available:~\url{http://www.cbsr.ia.ac.cn/english/CASIA-WebFace-Database.html}.

\bibitem{zhang1999shape}
Ruo Zhang, Ping-Sing Tsai, James~Edwin Cryer, and Mubarak Shah.
\newblock Shape-from-shading: a survey.
\newblock {\em IEEE Transactions on Pattern Analysis and Machine Intelligence},
  pages 690--706, 1999.

\bibitem{zhang2008cat}
Weiwei Zhang, Jian Sun, and Xiaoou Tang.
\newblock Cat head detection-how to effectively exploit shape and texture
  features.
\newblock In {\em Proceedings of the European Conference on Computer Vision},
  2008.

\bibitem{zhou2017unsupervised}
Tinghui Zhou, Matthew Brown, Noah Snavely, and David~G Lowe.
\newblock Unsupervised learning of depth and ego-motion from video.
\newblock In {\em Proceedings of the {IEEE} Conference on Computer Vision and
  Pattern Recognition}, 2017.

\bibitem{zhu2017rethinking}
Rui Zhu, Hamed Kiani~Galoogahi, Chaoyang Wang, and Simon Lucey.
\newblock Rethinking reprojection: Closing the loop for pose-aware shape
  reconstruction from a single image.
\newblock In {\em Proceedings of the International Conference on Computer
  Vision}, 2017.

\bibitem{Zhu2016Face}
Xiangyu Zhu, Zhen Lei, Xiaoming Liu, Hailin Shi, and Stan~Z. Li.
\newblock Face alignment across large poses: A {3D} solution.
\newblock In {\em Proceedings of the {IEEE} Conference on Computer Vision and
  Pattern Recognition}, 2016.

\bibitem{zimmermann2017learning}
Christian Zimmermann and Thomas Brox.
\newblock Learning to estimate 3d hand pose from single rgb images.
\newblock In {\em Proceedings of the International Conference on Computer
  Vision}, 2017.

\end{thebibliography}
}

\end{document}